\documentclass[10pt,letterpaper]{article}

\usepackage{ccn}
\usepackage{pslatex}
\usepackage{apacite}

\usepackage{color,soul}
\usepackage{amsmath,amsthm,amsfonts,amssymb}
\usepackage{enumerate}

\newcommand{\zset}[1]{\big\{ #1 \big\}}
\newcommand{\oo}[2]{#1{\tiny $\pm$ #2}}
\newcommand{\bo}[2]{\textbf{#1}{\tiny $\pm$ \textbf{#2}}}

\title{Synthesizing Deep Neural Network Architectures using Biological Synaptic Strength Distributions}

\author{{\large \bf A. H. Karimi (a6karimi@uwaterloo.ca)$^1$, \large \bf M. J. Shafiee (mjshafiee@uwaterloo.ca)$^2$ }\\
	 {\large \bf A. Ghodsi (aghodsib@uwaterloo.ca)$^3$, \large \bf A. Wong (a28wong@uwaterloo.ca)$^2$ } \\
  $^1$Computer Science Department, $^2$Systems Design Engineering Department, $^3$Statistics \& Actuarial Sciences Department, \\
  University of Waterloo, Waterloo, Ontario, Canada}

\begin{document}

\maketitle

\begin{quote}
\end{quote}

\section{Introduction}

We have recently witnessed an explosive growth in machine learning research focused on modelling and real-world inference problems. Notably, deep learning models such as deep neural networks (DNN) are a particularly powerful and biologically inspired class of learning algorithms that have consistently demonstrated state-of-the-art performance on tasks such as object recognition, image classification, image segmentation, and speech recognition. A particular type of DNN that has proven to be very effective in recent year are convolutional neural networks (CNNs) (see \cite{hubel1968receptive}) which are architecturally made up of layers of neurons modelled after simple and complex cells in the visual cortex.

In order to train a DNN for a task such as classification, the synaptic strengths of the network are optimized based on training data. Optimizing a large-scale artificial neural architecture such as a CNN for classification in a generalizable manner, however, requires on a large number of input image samples. This may be prohibitive in many practical scenarios where labeled data is limited. To ameliorate this dependence, we explore whether it is possible to sidestep the training of a large portion of learnable parameters\textemdash synaptic strengths\textemdash in a neural network. More particularly, we are motivated by~\cite{eliasmith2012large} where strong modelling and inference performance was exhibited when random synaptic strengths are leveraged in modelling of functional brain computationally. This suggests that the inherent structure of deep neural networks may itself be enough to elicit a powerful modelling and inference performance even when the formation of synaptic strengths are random.

In particular, we draw inspiration from a number of studies that investigated the distribution of synaptic strengths in the biological brain. For example, it has been observed that the synaptic strengths of certain synapses such as the excitatory synapses can be well modelled as random variables following well-known distributions such as truncated Gaussians \cite{barbour2007can}. Furthermore, Song {\it et al.}~\cite{song2005highly} found that the underlying synaptic strengths follows  a log-normal distributions. Other studies~ \cite{martinez2003complex, cheong2013cortical} suggested a correlated relationship between synaptic  strengths in earlier layers of the visual cortex, specifically circular concentric receptive fields modelled after Lateral Geniculate (LGN) cells.

Inspired by the aforementioned observations~\cite{song2005highly,martinez2003complex, cheong2013cortical}, we perform an exploratory study on different uncorrelated and correlated probabilistic generative models for synaptic strength formation in deep neural networks and the potential influence of different distributions on modelling performance particularly for the scenario associated with small data sets.

\section{Methodology}

\begin{table*}
  \caption{Impact of different probabilistic generative models for synaptic strength generation on modelling performance for 3 small datasets (see text on how datasets were generated). The synaptic strengths of the convolutional layers were generated from distributions describing synaptic strengths in the visual cortex. The convolutional layer synapses are frozen and not trained, whereas the fully connected layers of the CNN are trained over. Highest performing setups are in bold.
  }
  \label{results_summary}
  \centering
  \begin{tabular}{c|cccc}
    \hline
    Dataset  & Normal                      & Log-Normal                  & Center-Surround             & Fully Trained              \\

    \hline

    CIFAR-10 & \oo{19.83}{00.74}           & \oo{25.67}{00.17}           & \bo{26.40}{00.74}           & \oo{20.37}{00.62}          \\ 
    MNIST    & \oo{72.52}{00.21}           & \bo{80.89}{00.45}           & \oo{78.08}{01.19}           & \oo{79.01}{01.39}          \\ 
    SVHN     & \oo{26.40}{00.22}           & \bo{30.32}{00.17}           & \bo{30.86}{00.57}           & \oo{27.70}{01.79}          \\ 

    \hline
  \end{tabular}
  \vspace{-5mm}
\end{table*}

Here we model the synaptic strength distribution of the deep neural network as $P(\mathcal{W})$ where $\mathcal{W}$ is the set of synaptic strengths $\mathcal{W} = \zset{w_i}_1^n$ and $n$ is the number of synapses. In order to explore the effect of different probabilistic generative models for synaptic formation on modelling and inference performance in a focused manner, in this study we restrict the network architecture to be a convolutional neural network (CNN) architecture. More specifically, the synaptic strengths in the convolutional layers are synthesized based on $P(\mathcal{W})$ and are not fine-tuned, whereas the synaptic strengths of fully connected layers are synthesized and then trained to reach to their complete modelling capabilities. This setup allows us to localize the effect of $P(\mathcal{W})$ on synaptic strengths and fairly compare the modelling and inference performance of different synaptic formation drawn from various underlying biologically-inspired probability distributions. Furthermore, each random variable corresponding to a synaptic strength denoted as $w_i$ are drawn from a probabilistic generative model $P(\mathcal{W})$. In this study, we explore three different distribution models based on past biological studies:

\begin{enumerate}[I]
  \itemsep0em
  \item Normal Gaussian: $ P(\mathcal{W}) = \prod_{i = 1}^n \frac{1}{\sqrt{2 \pi}} \exp( - w_i^2 / 2 ) $
  \item Log-normal: $ P(\mathcal{W}) = \prod_{i = 1}^n \frac{1}{w_i \sigma \sqrt{2 \pi}} \exp \Big( \frac{\ln(w_i - \mu) ^ 2}{2 \sigma ^ 2} \Big) $
  \item Correlated center-surround: \\ $ P(\mathcal{W}) = \prod_{i = 1}^n \frac{1}{\sqrt{|2 \pi \Sigma_i|}} \exp( - \frac{w_i^T \Sigma_i w_i}{2} ) $\footnotemark
\end{enumerate}

\footnotetext{$\Sigma_i$ is the covariance matrix at synapse $i$, where the non-zero off-diagonal elements characterize the correlation between neighboring synapses.}

This approach to synapse strength formation can enable a drastic reduction in the number of parameters that need to be trained, which is an important factor in scenarios with small number of training data.

\subsection{Experimental Setup}

Followed by biological observations, the effect of three different $P(\mathcal{W})$ are examined on a same convolutional neural network (CNN) architecture here: I) normal Gaussian distribution, II) log-normal Gaussian distribution (\mbox{$\mu = - 0.702$}, \mbox{$\sigma^2 = 0.9355$} from \cite{song2005highly}), and III) correlated center-surround distribution.

In order to experiment the effect of different synaptic strength distributions on modelling performance, a CNN is utilized consisting of a convolutional layer comprising of 64 kernels with receptive fields of size $5 \times 5$, a max-pooling layer with stride 2, and a rectified non-linear unit, as well as two fully connected layers inspired by LeNet's fully connected layer architecture \cite{MNIST} and have a $1024N - 64N - 10N$ structure (input - hidden - output).

In this exploratory study, we examined three standard and publicly available object classification datasets including MNIST hand-written digits \cite{MNIST}, Street View House Numbers SVHN \cite{SVHN}, and CIFAR-10 object recognition dataset \cite{CIFAR-10} for the scenario of small training datasets. To mimic such a scenario $38$ samples per each class label (i.e., $10$ class labels for each dataset) were randomly selected from the available training data in each dataset to form a small dataset. However to compute the test accuracy, the models are tested with  all available testing samples. The reported results (mean and std) are computed based on three runs.

\vspace{-1.5mm}
\section{Preliminary Results}
Table \ref{results_summary} summarizes the results of our experiments. We also report the classification performance of the same CNN architecture on these datasets where the CNN is completely trained, and all synaptic strengths are fine-tuned. As expected, the small number of training samples (i.e., $38$ per class) results in the CNN's relatively poor classification performance, as is evident from the right-most column of Table \ref{results_summary} named ``Fully Trained''.

Interestingly, sampling the convolutional synaptic strengths from a normal Gaussian distribution (``Normal'' column) yields a classification performance comparable to that of ``Fully Trained'' for CIFAR-10 and SVHN. The most surprising of the preliminary results can be seen in the ``Log-Normal'' and ``Center-Surround'' columns. One possibility that these results suggest is that sampling the synaptic strengths of a CNN from well-known distributions that model synaptic strengths in the visual cortex can result in a classification system that potentially outperforms carefully fine-tuned CNNs on small datasets.  This may suggest that in the scenario with very little data, learning a generalizable classification system may not be worth the effort put into training as the performance may be outperformed by random convolutional synaptic strengths.  This result is a powerful first step towards designing deep neural networks that do not require many data samples to learn, and can sidestep / reduce the burden of current training procedures while maintaining or boosting classification and modelling performance. In future work, we are excited to explore this same effect on deeper networks with more synapses, and to investigate how and whether these synaptic strength distributions may be used to design more efficient architectures and training algorithms.

\vspace{-1.5mm}
\section{Acknowledgments}
This work was supported by the Natural Sciences and Engineering Research Council of Canada, Ontario Ministry of Economic Development and Innovation and Canada Research Chairs Program. The authors also thank Nvidia for the GPU hardware used in this study through the Nvidia Hardware Grant Program.

\vspace{-4mm}
\bibliographystyle{apacite}

\setlength{\bibleftmargin}{.125in}
\setlength{\bibindent}{-\bibleftmargin}

 \bibliography{refs}

\end{document}